# Preparation of Sentiment tagged Parallel Corpus and Testing its effect on Machine Translation

Sainik Kumar Mahata[1,] Amrita Chandra[2], Dipankar Das[3], Sivaji Bandyopadhyay[4]

[1,2,3,4] Jadavpur University, Kolkata, India
[1]sainik.mahata@gmail.com, [2]amritac03@gmail.com,
[3]dipankar.dipnil2005@gmail.com, [4]sivaji_cse_ju@yahoo.com

**Abstract.** In the current work, we explore the enrichment in the machine translation output when the training parallel corpus is augmented with the introduction of sentiment analysis. The paper discusses the preparation of the same sentiment tagged English-Bengali parallel corpus. The preparation of raw parallel corpus, sentiment analysis of the sentences and the training of a Character Based Neural Machine Translation model using the same has been discussed extensively in this paper. The output of the translation model has been compared with a base-line translation model using automated metrics such as BLEU and TER as well as manually.

## 1    Introduction

**M**achine **T**ranslation (MT) is the translation of one language to another with the help of computer software. MT is a difficult task and involves a thorough understanding of the source as well as the target text [8].

To train a good MT model, a large and good quality Parallel Corpus is required [13], where a parallel corpus is a collection of bilingual translated texts 14]. In simple words, if two languages are involved: the source monolingual text is an exact translation of the target monolingual text. Unfortunately, these resources are often scarce, limited in size, and have limited language coverage.

Since, MT is phenomena by which, in semantic level, machine translates one language to another, the translation quality takes a hit as state-of-art approaches don't dwell into the pragmatic level when translating. Our idea was that of introducing pragmatic features into MT, such that it improves the quality of translation.

Sentiments express the attitude and emotional condition of the speaker. So, it plays a major role in MT [10]. Since, a parallel corpus has a major impact on **S**tatistical **Ma**chine **T**ranslation (SMT) and **N**eural **M**achine **T**ranslation (NMT), the performance of

2MT can be increased if sentiment, by any means, can be added to the parallel corpus. For this, we have prepared a sentiment tagged parallel corpus for English-Bengali language pair. Furthermore, we compare the effectiveness of this corpus with a character based NMT model trained using a non-sentiment tagged parallel corpus, using metrics such as BLEU [11] and **T**ranslation **E**rror **R**ate (TER) [16] and manual evaluation.

Also, it has been established that MT systems work better when trained using simple sentences only [9]. In that event, we wanted to check whether this holds true for sentiment enriched simple sentences as well. Consequently, we prepared two additional sentiment tagged parallel corpora; one comprising of only simple sentences and another, consisting of "Other" (complex and compound) sentences.

The paper has been organized as follows. Section 2 describes a brief survey about the work done in this domain so far. Section 3 describes the methodology of text simplification, preparation of the sentiment tagged parallel corpus and training a character based NMT model with the same. Section 4 will discuss the results and will show the comparison of the created model with the baseline models. This will be followed by the Conclusion and Future Scope in Section 5.

## 2    Related Work

Various works have already been done on text simplification, sentiment analysis and generation of parallel corpora. Petersen et. al. [12] worked on text simplification for language learners. Simplified texts are commonly used by teachers and students in bilingual education and other language-learning contexts.

Claire Cardie et. al. [3] found out that finding simple, non-recursive, base noun phrases are an important subtask in many **N**atural **L**anguage **P**rocessing (NLP) applications. Kerstin Denecke [6] introduced a methodology for determining the polarity of text within a multilingual framework. The method leveraged on lexical resources for sentiment analysis available in English SentiWordNet[1]

Aurangzeb Khan et. al. [7] proposed the rule-based domain-independent sentiment analysis method. The proposed method classified subjective and objective sentences from reviews and blog comments. Federico Zanettin [20] worked on how small bilingual corpora of either general or specialized language can be used to devise a variety of structured and self-centered classroom activities whose aim was to enhance the understanding of the source language text and the ability to produce fluent target language texts.

Colin Bannard et. al. [2] worked on Using alignment techniques from phrase-based statistical machine translation, they showed how paraphrases in one language can be identified using a phrase in another language as a pivot. Daniel Varga et. al. [18] worked on a general methodology for rapidly collecting, building, and aligning parallel corpora for medium density languages, illustrating their main points on the case of Hungarian, Romanian, and Slovenian.

---

[1] http://sentiwordnet.isti.cnr.it/

Philip Resnik et. al. [15] worked on using the STRAND [14, 13] system for mining parallel text on the WorldWideWeb (WWW). Stefano Baccianella et. al. [1] worked on presenting SENTIWORDNET 3.0, a lexical resource explicitly devised for supporting sentiment classification and opinion mining applications.

Santanu Pal et. al. [10] worked on how sentiment analysis can improve the translation quality by incorporating the roles of sentiment holders, sentiment expressions and their corresponding objects and relations.

No work has been done so far, with respect to the creation of a sentiment tagged parallel corpus. On top of that, creating the same corpus with sentences of various complexities, have not been explored as well.

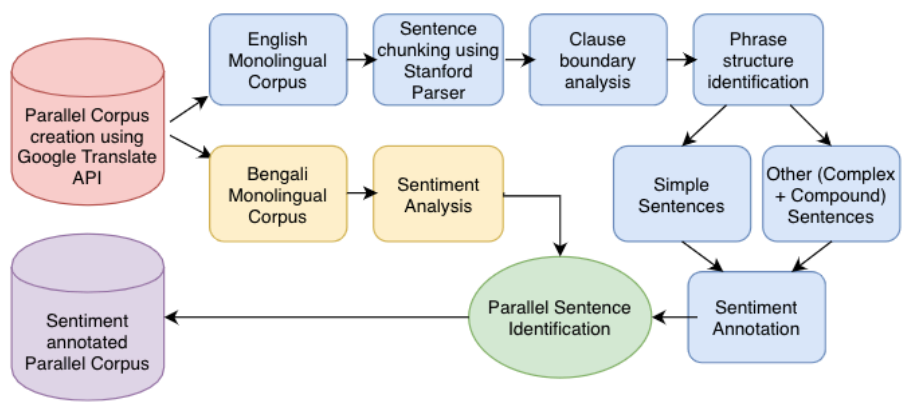

**Fig. 1.** Steps for preparation of the sentiment tagged parallel corpus.

## 3 Methodology

Our method involves the collection of English sentences initially and translating them into its corresponding Bengali counterpart to create a parallel corpus. For the additional resource involving simple sentences, the English data was shallow parsed to get the phrase structures. Using these, structure, the sentences were divided into simple and other sentences. Later, sentiment annotation for both the English and Bengali sentences were done, using various lexicons and this led to the preparation of the sentiment tagged parallel corpus. This method is shown in Figure 1. All the steps are discussed in details in the following subsections.

### 3.1 Collection of English sentences

We had an English-Bengali parallel corpus from **T**echnology **D**evelopment for **I**ndian **L**anguages Programme (TDIL)[2], which consisted of 49,999 parallel sentence pairs. In addition to this, we collected 57,985 English sentences from the resource of **M**achine **T**ranslation in **I**ndian **L**anguages (MTIL) shared task[3], organized by Amrita University and the statistics are shown in Table 1.

**Table 1.** Statistics of collected English sentences.

| Source | Data Size |
|---|---|
| Amrita University | 57,985 |
| TDIL | 49,999 |
| Total | 1,07,984 |

### 3.2 Translation using Google Translate API

The extracted English sentences, apart from the one provided by TDIL, were translated into Bengali using the Google Translate API[4] for Python. Then the English sentences and their Bengali translations were merged to obtain an additional parallel corpus. Asa result, the total number of English-Bengali parallel sentence pairs extracted for this experiment was 1,07,984. This corpus has been termed as "General Corpus" for the rest of the paper.

### 3.3 Shallow Parsing

Apart from the preparation of the sentiment tagged parallel corpus, we also wanted to partition the prepared corpus into **I.** simple sentences only and **II.** Other (Complex and Compound) sentences only. This is due to the fact that MT systems work better when trained using simple sentences. So, we decided to create parallel corpora, pertaining to different complexities, enriched with sentiment features as well.

To accomplish the classification, shallow parsing was used. Shallow Parsing is an analysis of a sentence in which constituent parts of sentences (nouns, verbs, adjectives, etc.) are identified and then higher order units that have discrete grammatical meanings (noun groups or phrases, verb groups, etc.) are linked. We have used Stanford Parser[5] in conjunction with Natural Language Toolkit[6] (NLTK) for performing shallow parsing on the English sentences. We avoided shallow parsing the Bengali sentences as no standard library were available for the same. An example of shallow parsing is shown in Table 2. Hereafter, clause identification of the English sentences was done as discussed in Section 3.4.

---

[2] http://tdil.meity.gov.in/
[3] http://nlp.amrita.edu/mtil_cen/
[4] https://pypi.org/project/googletrans/
[5] https://nlp.stanford.edu/software/lex-parser.shtml
[6] http://www.nltk.org/

Table 2. Statistics of collected English sentences.

| Sentence before Parsing |
|---|
| Initially, it ran on 6 routes which joined most of Delhi's parts. |
| **Sentence after Parsing** |
| S (ADVP (RB initially)) (, ,) (NP (PRP it)) (VP (VBD ran) (PP (IN on)(NP (NP (CD 6) (NNS routes)) (SBAR (WHNP (WDT which)) (S (VP (VBDjoined) (NP (NP (JJS most)) (PP (IN of) (NP (NP (NNP Delhi) (POS 's))(NNS parts)))))))))) (. .)) |

### 3.4 Clause Identification

**Identification of Simple Sentences**

A simple sentence is defined as a sentence which contains only one independent clause and has no dependent clauses. Generally, when-ever two or more clauses are joined by conjunctions (coordinating and subordinating), it becomes a complex or a compound sentence accordingly. So, to get a hold on handling the conjunctions, we used the Stanford parser to shallow parse the English sentences into phrases. (viz.NP (Noun Phrase), VP (Verb Phrase), PP (Preposition Phrase), ADJP (Adjective Phrase) and ADVP (Adverb Phrase)).

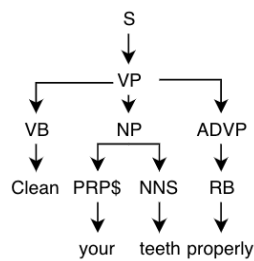

**Fig. 2.** Extraction of phrase chunks.

We noticed that simple sentences have a unique phrase structure that consists of combinations of NP, VP, and PP as shown in Figure 2. In conjunction with this theory, we applied a rule-based approach to extract simple sentences from the English corpus. We subjected 3,046 simple sentences to shallow parsing and extracted the unique phrase structures. This constituted the rules by which we further mined for simple sentences from the English corpus. We extracted 205 unique rules, the surface forms of which, along with their confidence score, are shown in Table 3. Confidence Score was calculated as a fraction of the total number of sentences identified using a specific rule, by the total number of sentences, as denoted in equation 1.

$$\text{Confidence Score} = \frac{\text{No. of sentences identified using rule}}{\text{Total no. of sentences}} \quad (1)$$

Table 3. Surface forms of the extracted rules. "*" means one or more occurrence of item.

| Rules | Confidence |
|---|---|
| PP NP* PP VP NP* | 8.40 |
| PP NP* VP PP NP* | 9.49 |
| ADVP NP* VP* ADVP NP* | 9.36 |
| NP VP PP NP PP NP | 12.15 |
| NP ADVP VP* NP* | 11.69 |
| NP* VP NP* | 11.69 |
| NP* PP NP VP* NP | 11.46 |
| NP VP PP NP* | 11.23 |
| VP* NP* PRP* ADVP* | 4.92 |
| NP VP* NP* PP* ADJP* ADVP* | 9.62 |

We tested our system on 2,876 sentences (1,438 simple sentences and 1438 complex/compound sentences) and got an accuracy of 89.22%. Table 4 shows the confusion matrix for the same. We used this system to extract 16,654 simple sentences from the general corpus.

Table 4. Confusion matrix for the rule based approach followed to extract simple sentences.

|  | Other | Simple | Kappa |
|---|---|---|---|
| Other | 1275 | 90 |  |
| Simple | 220 | 1291 |  |
| Prec. | 93.41% |  | 0.78 |
| Recall | 85.28% |  |  |
| Acc. | 89.22% |  |  |
| F1 | 89.16% |  |  |

**Identification of Other sentences**
*Complex*
POS tags assigned to every token in the shallow parsed output, were used to identify the complex sentences. Rules followed for extracting complex sentences are as follows, where SBAR tags denotes a subordinating conjunction.

- If a line in a shallow parsed file contains SBAR tag in between two sentences then we can say that the sentence is a complex sentence.
- If a line in a chunked file contains SBAR in starting of the sentence and ',' in middle of the sentence then the sentence is considered as a complex sentence.

An example of a phrase structure of a complex sentence is shown in Figure 3. Using these rules, we were able to extract 39,068 complex sentences from the general corpus.

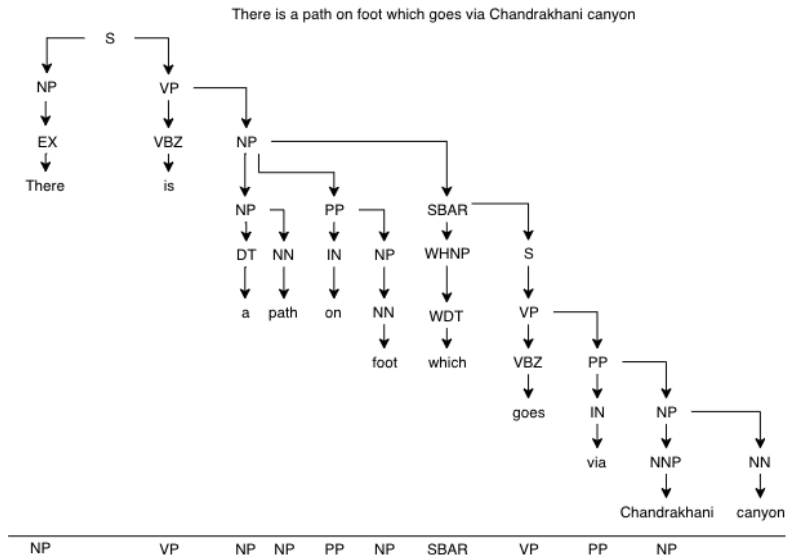

**Fig. 3.** Example of complex sentence phrase structure.

*Compound*

Similarly, we identified rules for extracting compound sentences as well. If the shallow parsed sentence has a CC (coordinating conjunction) tag followed by S (starting of a sentence) tag, then the sentence is a compound sentence. Symbolically we can define the rule as follows.

$$S_{Start} \ldots \text{Other POS} \ldots CC_{POS} \longrightarrow S_{Start} \ldots \text{Other POS}$$

An example of phrase structure of a compound sentence is shown in Figure 4.

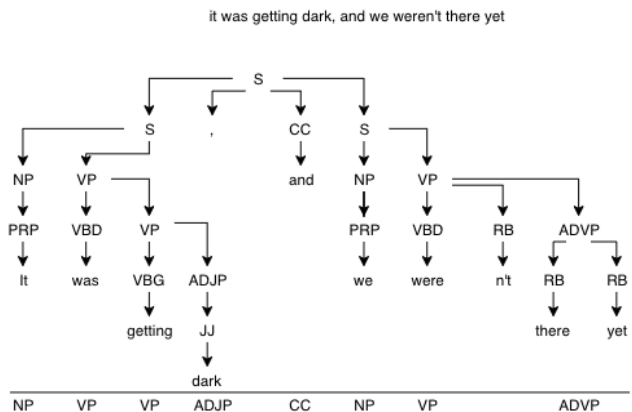

**Fig. 4.** Example of compound sentence phrase structure.

Using these rules, we were able to extract 43,703 complex sentences from the general corpus. The number of simple, complex and compound sentences extracted with the help of Section 3.4 are shown in Table 5. Our system could not tag 8,559 sentences into any category. It is to be noted that, our hypothesis considers that the Bengali sentences have the same complexity, like that of its English counterpart.

Table 5. No. of sentences extracted using rule based approach.

| Type | | Number of Sentences |
|---|---|---|
| Simple | | 16,654 |
| Others | Complex | 39,068 |
| | Compound | 43,703 |
| Untagged | | 8,559 |

### 3.5 Sentiment Annotation

**S**entiment **A**nalysis (SA) refers to the use of NLP to systematically identify, extract, quantify, and study affective states and subjective information. In this work, we have tried to use sentiment annotation in parallel corpus to help increase the performance of MT. This was done by tagging the English words in a sentence $S_e$ and Bengali words in a sentence $S_b$, with their sentiments, by using SentiWordNets of the respective languages. Here, $S_e$ and $S_b$ are parallel sentences in the general corpus. This process created sentiment tagged English and Bengali sentences $S^{st}_e$ and $S^{st}_b$ respectively. Consequently, these sentences were termed parallel, if these satisfied certain rules, that has been described in Section 3.5. Also, sentiment tagging of the English and Bengali words has been described in Section 3.5. This experiment was done to study whether enriching parallel corpus with sentiment features enhance the performance of MT or not. Also, since these types of corpora are hard to find, creating one will aid in future research.

**Word Level Sentiment Tagging**
SentiWordNet of positive and negative words for English[7] and Bengali[8] were applied to the English and Bengali files separately. We also used additional lexicons like AFINN-96[9], AFINN-111[10], Taboada Grieve 2004-SO [17] and Vender Sentiment, for English. This step was repeated for the general corpus, Simple sentence corpus, and the "Other" (Complex, Compound) sentences. It is to be noted that positive sentiment and negative sentiment were annotated as POS and NEG respectively. The neutral sentiments were not annotated. A snippet of the result of this step is shown in Table 6. The result of this step is the generation of sentiment tagged English and Bengali sentences, $S^{st}_e$ and $S^{st}_b$, respectively.

---

[7] http://www.nltk.org/howto/sentiwordnet.html
[8] http://amitavadas.com/sentiwordnet.php
[9] http://www2.imm.dtu.dk/pubdb/views/publication_details.php?id=5981
[10] https://udel.instructure.com/courses/1310886/files/57548375

Table 6. Example of sentiment annotation in both English and Bengali.

| English sentiment annotated sentence | Bengali sentiment annotated sentence |
|---|---|
| The tourists admired\POS the paintings. | পর্যটকরা পেইন্টিংসকে প্রশংসিত\POS করেছেন। |
| I admired\POS him for his honesty. | আমি তার সততা\POS জন্য তাকে প্রশংসিত। |
| The enemy\NEG soldiers submitted to us. | শত্রু\POS\NEG সৈন্য আমাদের জমা\POS\NEG দেওয়া। |
| Ellen warned\NEG Helen that the party\POS would be tonight. | এলেন হেলেন কে সতর্ক\POS\NEG করে\NEG দিয়ে বলেন পার্টি আজ রাতে হবে। |

**Sentiment Tagged Parallel Corpus**

Sentiment tagged English and Bengali sentences, $S^{st}_e$ and $S^{st}_b$ were considered parallel if they satisfied the following rules.

R1. If the English sentence $S^{st}_e$ is having one or more POS tag and its corresponding Bengali sentence $S^{st}_b$ is also having one or more POS tag, they are considered as parallel.

R2. If the English sentence $S^{st}_e$ is having one or more NEG tag and its corresponding Bengali sentence $S^{st}_b$ is also having one or more NEG tag, they are considered as parallel.

R3. If the English sentence $S^{st}_e$ is having one or more POS and NEG tag and its corresponding Bengali sentence $S^{st}_b$ is also having one or more POS and NEG tag, they are considered as parallel.

R4. If the English sentence $S^{st}_e$ is having one or more POS and NEG tag and its corresponding Bengali sentence $S^{st}_b$ is also having one or more POS tag, and vice versa, they are considered as parallel.

R5. If the English sentence $S^{st}_e$ is having one or more POS and NEG tag and its corresponding Bengali sentence $S^{st}_b$ is also having one or more NEG tag, and vice versa, they are considered as parallel.

It is to be noted that, sentences with no POS and NEG tags were not considered as it did not solve the purpose of enriching the sentiment tagged parallel corpus. These steps were followed for the general, simple sentence and the "Other" sentence corpus. The statistics of the above is given in Table 7.

Table 7. Statistics of extracted Sentiment tagged parallel corpus.

| Corpus | No. of Sentiment tagged parallel sentences |
|---|---|
| General | 70,357 |
| Simple | 6,700 |
| Others | 63,619 |

### 3.6 Character based Neural Machine Translation

The Neural Machine Translation model (sequence to sequence model) is a relatively new idea for sequence learning using neural networks. It has gained quite some popularity since it achieved state of the art results in machine translation task. Essentially, the model takes as input a sequence

$$X = \{x_1, x_2, \ldots, x_n\}$$

and tries to generate the target sequence as output

$$Y = \{y_1, y_2, \ldots, y_n\},$$

where $x_i$ and $y_i$ are the input and target symbols respectively. The architecture of seq2seq model comprises of two parts, the encoder and decoder. Since we wanted to test our model at character level, the inputs to the encoder and decoder were characters of the source and target sentences. We implemented both the models using the Keras [4] library. Character level NMT (CNMT) performs better than Word Level NMT due to the following reasons

- It does not suffer from out-of-vocabulary issues
- It is able to model different, rare morphological variants of a word
- It does not require segmentation [5].

Generally, CNMT works best when majority of alphabets, in the source and target language, overlap i.e., both the languages share a common or similar script. Still, we tried to find out its performance on the simple sentence and whole corpus, though in our case Devanagari script and Roman script utilizes completely different alphabets. The model has two parts (encoder and decoder) as discussed below.

**Encoder** For building the encoder we used an embedding layer and two **L**ong **S**hort **T**erm **M**emory (LSTM) layers. An embedding layer turns positive integers into dense vectors of fixed size. One hot tensors of the English sentences (embeddings at character level) were fed to the input layer. Thereafter, the embedding layer converted these into dense vectors. From the encoder, the internal states of each cell were preserved and the outputs were discarded. The purpose of this is to preserve the information at context level. These states were then passed on to the decoder cell as initial states.

**Decoder** For building the decoder, again an LSTM layer was used with initial states as the hidden states from encoder. It was then fed to another LSTM layer. It was designed to return both sequences and states. Here, the concept of teacher forcing [19] was used. The input to the decoder was one hot tensor (embeddings at character level) of Bengali sentences while the target data was identical, but with an offset of one time-step ahead. This was fed to an embedding layer and then to a LSTM layer. The information for generation is gathered from the initial states passed on by the encoder. Thus, the de-coder learns to generate target data [t+1, ...] given targets [..., t] conditioned on the input sequence. It essentially predicts the output sequence, one character per output time step. For training the model, various sets of English and Bengali sentences were fed to the CNMT model. These sets will be discussed in Section 4. The batch size was set to64, number of epochs was set to 100, activation function was softmax, optimizer chosen was rmsprop and loss function used was categorical cross-entropy. Learning rate was set to 0.001. The general architecture of the created model is shown in Figure 5.

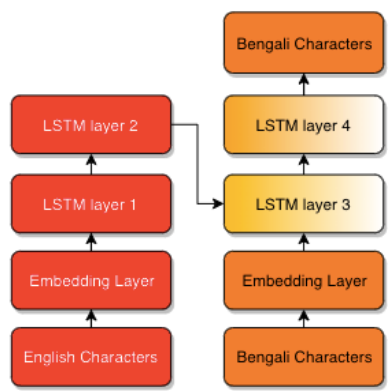

**Fig. 5.** Architecture of the character embedded NMT model.

## 4    Results

Three sets of experiments were performed with the prepared data. They are listed below.
- Comparison between the general corpus of size 1,07,984 and sentiment tagged general corpus of size 70,357, sentence pairs.
- Comparison between non-sentiment tagged simple sentence corpus of size 16,654and sentiment tagged simple sentence corpus of size 6,700, sentence pairs.
- Comparison between non-sentiment tagged other sentence corpus of size 82,771and sentiment tagged other sentence corpus of size 63,619, sentence pairs.

CNMT model was trained using the aforementioned data and automated and manual evaluation of the same is shown in Section 4.1.

### 4.1    Evaluation

To create the CNMT models, that would be used to test the effectiveness of the created sentiment tagged parallel corpus on MT, we have used a non-sentiment tagged parallel corpus as well as the sentiment tagged parallel corpus (General, Simple and Other). All the created models were tested using 200 sentences. The comparison between all the models, with respect to automated evaluation, is shown in Table 8.

Translation quality was judged by a linguist with Bengali as his mother tongue. The evaluation criteria were Adequacy and Fluency. Adequacy means how much of the meaning expressed in the target translation. Fluency means to what extent the translation is well-formed grammatically, contains correct spellings and intuitively acceptable and can be sensibly interpreted by a native speaker. The linguist was asked to rate the translation in the range of 1-5, where '1' is the lowest and '5' is the highest. The manual evaluation measures for the CNMT model, when trained using various datasets, is given in Table 9.

**Table 8.** Automated evaluation of the CNMT model when trained with various datasets.

| Type | Metric | General | Simple | Other |
|---|---|---|---|---|
| Non-sentiment | BLEU | 10.25 | 5.67 | 8.92 |
| tagged | TER | 81.01 | 92.14 | 89.27 |
| Sentiment | BLEU | 11.51 | 4.68 | 9.16 |
| tagged | TER | 78.59 | 95.47 | 87.54 |

**Table 9.** Manual evaluation of the CNMT model when trained with various datasets.

| Type | Metric | General | Simple | Other |
|---|---|---|---|---|
| Non sentiment | Adequacy | 2.98 | 1.91 | 2.76 |
| tagged | Fluency | 2.56 | 1.47 | 2.67 |
| Sentiment | Adequacy | 3.03 | 1.59 | 2.89 |
| tagged | Fluency | 2.84 | 1.62 | 2.73 |

## 5   Conclusion and Future Scope

In this paper, we have discussed the procedure we followed to build a sentiment tagged parallel corpus. Additionally, we have separately created two resources viz., a sentiment tagged parallel corpus with only simple sentences and another one with
"Other" sentences. We can clearly see from the automated and manual evaluation that an NMT model when trained using a sentiment tagged corpus, perform slightly better in both BLEU, and TER, when compared to the baseline system trained using non-sentiment tagged parallel corpus. We can also see from the manual evaluation that fluency improves moderately with the sentiment tagged corpus. Therefore, we can safely say, sentiment does play a role in improving MT output quality.

As a future work, we would like to experiment the same for SMT and Word level NMT. Moreover, since the tagging of sentiments has been done on a word level, using Senti-WordNets, we would like to find the sentiment of the entire sentence using a sentiment analysis system and representing the sentiment in the encoder representation. Also, we would like to check whether enriching parallel corpus with sentiment features leads to propagation of sentiment through the MT pipeline. If yes, this would greatly reduce post editing efforts as well.

**Acknowledgement**

This work is supported by Media Lab Asia, MeitY, Government of India, under the Visvesvaraya PhD Scheme for Electronics & IT.

**Reference**

1. Baccianella, S., Esuli, A., Sebastiani, F.: Sentiwordnet 3.0: An enhanced lexical resource for sentiment analysis and opinion mining. In: in Proc. of LREC (2010)
2. Bannard, C., Callison-Burch, C.: Paraphrasing with bilingual parallel corpora. In: Proceedings of the 43rd Annual Meeting on Association for Computational Linguistics, pp. 597–604.Association for Computational Linguistics (2005)
3. Cardie, C., Pierce, D.: Error-driven pruning of treebank grammars for base noun phrase identification. In: Proceedings of the 17th international conference on Computational Linguistics-Volume 1, pp. 218–224. Association for Computational Linguistics (1998)
4. Chollet, F., et al.: Keras. https://keras.io (2015)
5. Chung, J., Cho, K., Bengio, Y.: A character-level decoder without explicit segmentation for neural machine translation. CoRRabs/1603.06147(2016). URL http://arxiv.org/abs/1603.06147
6. Denecke, K.: Using sentiwordnet for multilingual sentiment analysis. In: Data Engineering Workshop, 2008. ICDEW 2008. IEEE 24th International Conference on, pp. 507–512. IEEE (2008)
7. Khan, A.: Sentiment classification by sentence level semantic orientation using sentiwordnet from online reviews and blogs. International Journal of Computer Science & EmergingTechnologies2(4) (2011)
8. Mahata, S.K., Das, D., Bandyopadhyay, S.: Mtil2017: Machine translation using recurrent neural network on statistical machine translation. Journal of Intelligent Systems pp. 1–7(2018)
9. Mahata, S.K., Mandal, S., Das, D., Bandyopadhyay, S.: SMT vs NMT: A comparison over Hindi & Bengali simple sentences. arXiv preprint arXiv:1812.04898 (2018)
10. Pal, S., Patra, B.G., Das, D., Naskar, S.K., Bandyopadhyay, S., Genabith, J.v.: How sentiment analysis can help machine translation. In: Proceedings of the 11th International Conference on Natural Language Processing, pp. 89–94. NLP Association of India (2014). URL http://www.aclweb.org/anthology/W14-5114
11. Papineni, K., Roukos, S., Ward, T., Zhu, W.J.: Bleu: a method for automatic evaluation of machine translation. In: Proceedings of the 40th annual meeting on association for computational linguistics, pp. 311–318. Association for Computational Linguistics (2002)

12. Petersen, S.E., Ostendorf, M.: Text simplification for language learners: a corpus analysis. In: Workshop on Speech and Language Technology in Education (2007)
13. Resnik, P.: Parallel strands: A preliminary investigation into mining the web for bilingual text. In: In Third Conference of the Association for Machine Translation in the Americas, pp. 72–82. Springer (1998)
14. Resnik, P.: Mining the web for bilingual text. In: In Proceedings of the 37th Annual Meeting of the Association for Computational Linguistics, pp. 527–534 (1999)
15. Resnik, P., Smith, N.A.: The web as a parallel corpus. Computational Linguistics29(3),349–380 (2003)
16. Snover, M., Dorr, B., Schwartz, R., Micciulla, L., Makhoul, J.: A study of translation edit rate with targeted human annotation. In: Proceedings of association for machine translation in the Americas, vol. 200 (2006)
17. Taboada, M., Grieve, J.: Analyzing appraisal automatically. In: In Proceedings of the AAAI Spring Symposium on Exploring Attitude and Affect in Text: Theories and Applications, pp.158–161 (2004)
18. Varga, D., Halácsy, P., Kornai, A., Nagy, V., Németh, L., Trón, V.: Parallel corpora for medium density languages. Amsterdam Studies in The Theory and History of Linguistic Science Series 4292, 247 (2007)
19. Williams, R.J., Zipser, D.: A learning algorithm for continually running fully recurrent neural networks (1989)20. Zanettin, F.: Bilingual comparable corpora and the training of translators. Meta: Journal destraducteurs/Meta: Translators' Journal43(4), 616–630 (1998)